\documentclass{article}

\usepackage[english]{babel}
\usepackage[utf8]{inputenc}
\usepackage[T1]{fontenc}
\usepackage{arxiv}
\usepackage{booktabs}
\usepackage{amsmath}
\usepackage{amsfonts}
\usepackage{microtype}
\usepackage{graphicx}
\usepackage{hyperref}
\usepackage{natbib}
\usepackage{doi}
\usepackage{xcolor}
\usepackage{orcidlink}
\usepackage{tikz}
\usepackage{algorithm}
\usepackage{algpseudocode}
\usepackage{listings}
\usepackage{pdfpages}
\usetikzlibrary{arrows.meta, positioning}
\AtBeginDocument{\setlength{\headheight}{23pt}}

\title{Automating Timed Up and Go Phase Segmentation and Gait Analysis via the tugturn Markerless 3D Pipeline}

\renewcommand{\shorttitle}{\textit{tugturn} Automated Markerless 3D TUG}

\author{
  Abel Gon\c{c}alves Chinaglia\orcidlink{0000-0002-6955-7187}$^{1,2}$ \\
  \texttt{abel.chinaglia@usp.br} \\
  \And
  Guilherme Manna Cesar\orcidlink{0000-0002-5596-9439}$^{3}$ \\
  \texttt{g.cesar@unf.edu} \\
  \And
  Paulo Roberto Pereira Santiago\orcidlink{0000-0002-9460-8847}$^{1,2}$ \\
  \texttt{paulosantiago@usp.br} \\
}

\date{
  $^1$ Biomechanics and Motor Control Laboratory, School of Physical Education and Sport of Ribeirao Preto, University of Sao Paulo, Brazil \\
  $^2$ Graduate Program in Rehabilitation and Functional Performance, Ribeirao Preto Medical School, University of Sao Paulo, Brazil \\
  $^3$ Department of Physical Therapy, Brooks College of Health, University of North Florida, USA \\
}

\hypersetup{
  pdftitle={tugturn: Markerless 3D TUG and Turn Analysis},
  pdfsubject={biomechanics, markerless motion analysis, timed up and go},
  pdfauthor={Abel Goncalves Chinaglia, Guilherme Manna Cesar, Paulo Roberto Pereira Santiago},
  pdfkeywords={Timed Up and Go, markerless motion capture, gait events, vector coding, XCoM, biomechanics},
  colorlinks=true,
  linkcolor=black,
  citecolor=black,
  urlcolor=blue
}

\begin{document}
\maketitle

\begin{abstract}
Instrumented Timed Up and Go (TUG) analysis can support clinical and research decision-making, but robust and reproducible markerless pipelines are still limited. We present \textit{tugturn.py}, a Python-based workflow for 3D markerless TUG processing that combines phase segmentation, gait-event detection, spatiotemporal metrics, intersegmental coordination, and dynamic stability analysis. The pipeline uses spatial thresholds to segment each trial into stand, first gait, turning, second gait, and sit phases, and applies a relative-distance strategy to detect heel-strike and toe-off events within valid gait windows. In addition to conventional kinematics, \textit{tugturn} provides Vector Coding outputs and Extrapolated Center of Mass (XCoM)-based metrics. The software is configured through TOML files and produces reproducible artifacts, including HTML reports, CSV tables, and quality-assurance visual outputs. A complete runnable example is provided with test data and command-line instructions. This manuscript describes the implementation, outputs, and reproducibility workflow of \textit{tugturn} as a focused software contribution for markerless biomechanical TUG analysis.
\end{abstract}

\keywords{Timed Up and Go \and markerless 3D analysis \and gait event detection \and Vector Coding \and XCoM \and biomechanics software}

\section{Introduction}
Timed Up and Go (TUG) is widely used to evaluate mobility and functional performance in clinical populations because it integrates postural transition, walking, turning, and sitting in a single task \citep{podsiadlo1991}. Traditional motion-capture workflows provide high-fidelity biomechanical information, but they often depend on specialized infrastructure and time-consuming processing pipelines. Markerless approaches based on computer vision have expanded access to movement analysis and enabled broader deployment in rehabilitation and applied biomechanics \citep{lugaresi2019mediapipe, nath2019using}.

Historically, gait events and turning metrics have been analyzed predominantly using standard optoelectronic kinematic systems, instrumented walkways, or force platforms \citep{khobkhun2022investigation, thanakamchokchai2025effects}. However, widely available full-body 3D-landmark frameworks custom-built for automated clinical phase segmentation remain scarce \citep{tahara2025predicting}. Without such specialized automated markerless tools, analyses in clinical or uncontrolled settings often resort to subjective visual recognition or labor-intensive manual frame-by-frame data processing. To overcome these limitations, the \textit{tugturn} method was developed to automate these demanding procedures, bridging the gap between raw video tracking and robust biomechanical endpoints.

Despite this progress, many markerless pipelines still emphasize either pose extraction or generic gait summaries and do not provide a structured TUG-specific workflow with robust phase-aware event handling. In TUG, meaningful interpretation depends on correctly isolating movement phases (e.g., first gait versus second gait) and on reducing false event detections outside valid locomotor windows. This issue is particularly relevant in heterogeneous clinical recordings, where pauses, partial turns, and variable trajectories are common.

The \textit{tugturn.py} module was designed to address this gap by integrating: (i) spatially constrained phase segmentation, (ii) gait-event detection based on relative-distance principles \citep{zeni2008}, (iii) phase-specific spatiotemporal metrics, (iv) Vector Coding for coordination analysis, and (v) Extrapolated Center of Mass (XCoM) indicators for dynamic stability assessment. The goal of this paper is to document the software architecture and reproducible processing flow of \textit{tugturn.py} as a practical tool for markerless 3D TUG analysis.

\section{Software Architecture and Methodology}
\begin{figure}[htbp]
  \centering
  \includegraphics[width=0.7\linewidth]{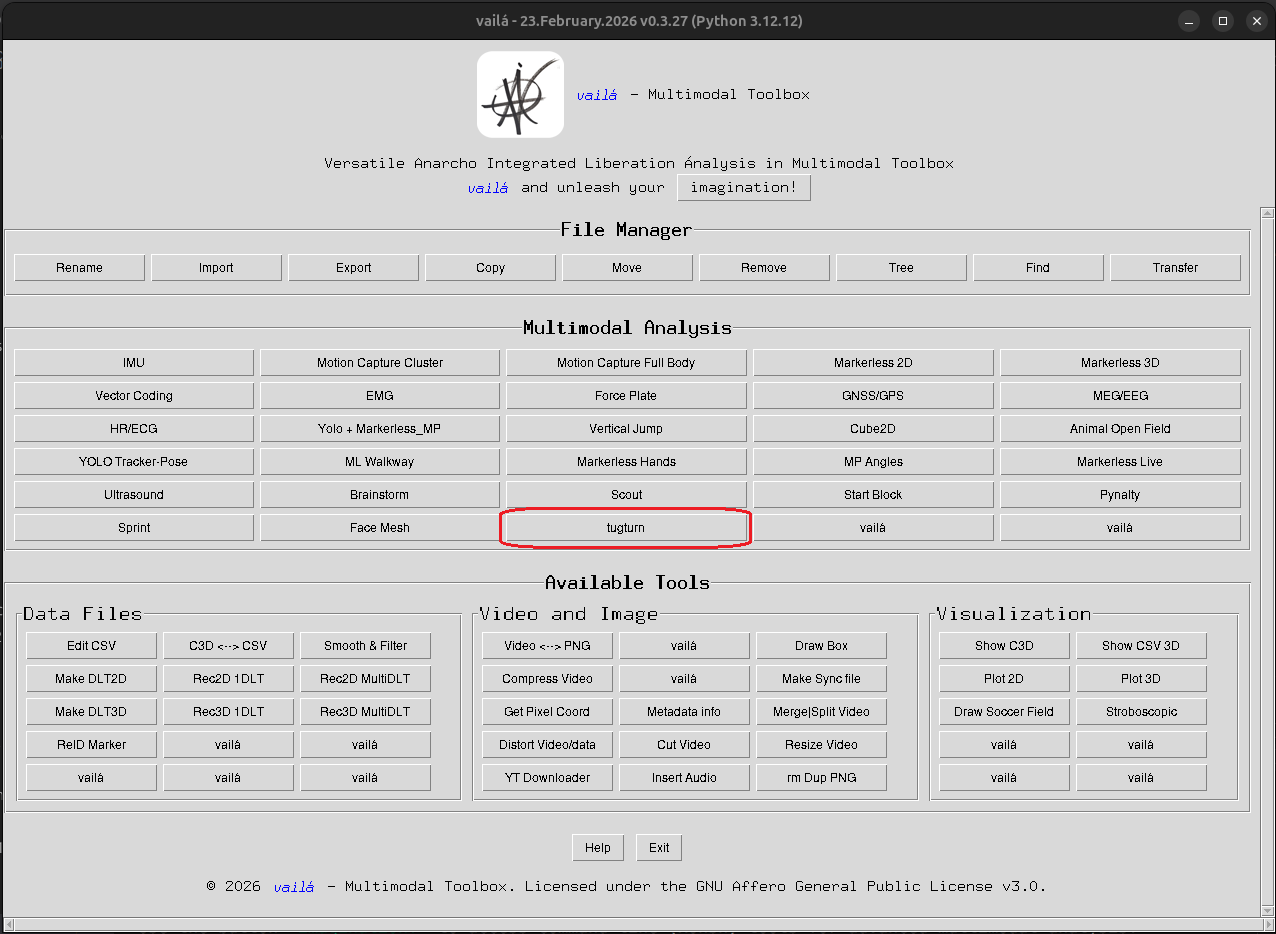}
  \caption{The \textit{vail\'{a}} GUI, highlighting the \textit{tugturn} button used to initialize the analysis module.}
  \label{fig:vaila_gui}
\end{figure}

The \textit{tugturn.py} module is designed as a highly cohesive command-line interface (CLI) driven tool. It ingests 3D coordinate time-series (typically derived from MediaPipe and processed via the \textit{vail\'{a}} \citep{vaila2024} framework) and executes a multi-stage deterministic pipeline. When \textit{vail\'{a}}  is used as a Graphical User Interface (GUI), the pipeline can be initiated by selecting the \textit{tugturn} button, as illustrated in Figure~\ref{fig:vaila_gui}.

\subsection{Spatial Phase Segmentation (The Y-Axis Logic)}
Traditional TUG assessments treat the movement as a temporal ``black box,'' providing only total completion time. To overcome this limitation and eliminate the ``blind stopwatch'' effect, \textit{tugturn.py} implements a deterministic spatial segmentation algorithm that tracks the estimated Center of Mass (pelvis) along the anteroposterior progression axis (Y-axis). By applying strict spatial thresholds --- specifically $Y \le 1.125$ m for the chair setup zone and $Y \approx 4.5$ m for the turning zone --- the pipeline automatically slices the continuous trial into five discrete functional sub-phases: Sit-to-Stand, First Gait, Turning, Second Gait, and Stand-to-Sit. This approach transforms a subjective duration test into a precise topographical examination of movement.

\subsection{Gait Event Detection (Dynamic Zeni Algorithm)}
Extracting gait events from a bidirectional test poses a significant challenge for classical coordinate-based algorithms, which often generate false positives during turns. To address this, we implemented a robust adaptation of the Zeni relative-distance algorithm \cite{zeni2008two}, enhanced by rigorous spatial and kinetic filters:

\begin{enumerate}
    \item \textbf{Walking Mask (Posture and Movement Filter):} Before step detection, the algorithm verifies that the subject is standing (pelvic Z-axis height within the upper third of its range) and moving (XY-plane velocity $> 0.15$ m/s). This prevents false detections during turning or postural transitions.
    \item \textbf{Dynamic Progression Vector:} Because the TUG involves both forward and return paths, we calculate the smoothed derivative of the pelvic center in the XY plane to generate a dynamic unit directional vector. This allows the detection to adapt frame-by-frame to the patient's actual walking direction.
    \item \textbf{Heel Strike (HS) Detection:} Following Zeni's logic, we calculate the vector distance between the heel and the pelvic center, project it onto the dynamic progression vector via dot product, and apply a peak detection function (\texttt{find\_peaks}) to find local maxima. A physiological constraint ensures no two consecutive strikes of the same foot occur within a 300 ms window.
    \item \textbf{Toe-Off (TO) Detection:} TO is identified by finding the local minima (negative peaks) of the toe marker's projection relative to the pelvis, representing the moment the trailing foot is maximally extended backwards.
    \item \textbf{Spatial Restriction (Double-Check):} Finally, detected frames are cross-referenced with the spatial segmentation zones. Any event falling outside the strict ``First Gait'' or ``Second Gait'' boundaries is systematically discarded.
\end{enumerate}

\subsection{Advanced Kinematics and Intersegmental Coordination}
Moving beyond basic spatiotemporal metrics, the \textit{tugturn.py} pipeline extracts deep biomechanical features through its Kinematics (KIN) module. This includes continuous monitoring of trunk inclination --vital for detecting postural abnormalities such as camptocormia --and lower limb joint excursions. Crucially, the system applies Vector Coding to quantify intersegmental coordination between the trunk and pelvis during the turning phase. By computing the continuous relative phase angle, the software categorizes turning behavior into In-Phase (indicative of Parkinsonian \textit{en bloc} turning) or Anti-Phase patterns. Previous Machine Learning implementations (XGBoost) on our clinical datasets have confirmed that these specific coordination metrics are among the top predictors of disease severity (e.g., modified Hoehn \& Yahr stages), establishing them as robust digital biomarkers.

\section{Implementation and Usage (High-Throughput Architecture)}
Traditional 3D biomechanical analysis often requires labor-intensive, frame-by-frame manual trajectory cleaning. In contrast, \textit{tugturn.py} is engineered for scalable, high-throughput batch processing. Written in pure Python and prioritizing minimal dependencies (NumPy, Pandas, SciPy, Matplotlib, Plotly), the software ensures cross-platform compatibility.

Driven by a command-line interface and configurable \texttt{.toml} files, the pipeline can automatically process entire clinical databases in minutes. Automated output generation includes detailed JSON datasets, kinematic CSV spreadsheets ready for statistical analysis (\texttt{bd\_kinematics}, \texttt{bd \_vector\_coding}), and rich interactive HTML reports with embedded GIF visualizations. This architecture successfully bridges the gap between complex data engineering and accessible clinical diagnostics.

\begin{lstlisting}[language=bash, caption=Example of CLI execution for a single trial or batch directory]
# Run analysis on a single trial with a spatial tolerance of 0.15m
python vaila/tugturn.py -i tests/s26_m1_t1.csv -c tests/s26_m1_t1.toml \
    -o results/ -y 0.15

# Batch process an entire directory of CSV files (e.g., hundreds of trials)
python vaila/tugturn.py -i data/raw_kinematics/ -c config/default.toml \
    -o data/processed/ -y 0.15
\end{lstlisting}

\section{Results-Oriented Software Deliverables}
Rather than presenting a new clinical trial in this manuscript, we report the software deliverables and processing scope demonstrated by the included test workflow. The pipeline provides:
\begin{itemize}
  \item deterministic phase segmentation with explicit spatial constraints,
  \item phase-aware HS/TO event extraction,
  \item kinematic and spatiotemporal summaries aligned with TUG sub-tasks,
  \item coordination and stability metrics in machine-readable tables,
  \item browser-ready reports for rapid quality control and interpretation.
\end{itemize}

Table~\ref{tab:deliverables} summarizes the core output families.

\begin{table}[htbp]
  \centering
  \caption{Core output families generated by \textit{tugturn.py}.}
  \label{tab:deliverables}
  \begin{tabular}{ll}
    \toprule
    \textbf{Output family} & \textbf{Purpose} \\
    \midrule
    HTML reports & Human-readable trial overview and plots \\
    \texttt{bd\_results} CSV & Global and phase metrics for analysis \\
    \texttt{bd\_steps} CSV & Step-level event and spatiotemporal values \\
    \texttt{bd\_kinematics} CSV & Time-series kinematic variables \\
    Vector Coding CSVs & Coupling-angle metrics and variability summaries \\
    Phase GIFs & Segmentation quality assurance \\
    \bottomrule
  \end{tabular}
\end{table}

\subsection{Example Run Output (HTML Reports)}
To demonstrate a real execution artifact in this manuscript, we include the reference run
\texttt{s26\_m1\_t1} generated in:
\texttt{tugturn/results/}. The corresponding HTML outputs are:
\begin{itemize}
  \item \url{tugturn/results/s26_m1_t1_tug_report.html}
  \item \url{tugturn/results/s26_m1_t1_tug_report_interactive.html}
\end{itemize}

From the same run, Table~\ref{tab:example_run_metrics} summarizes key values exported by
\texttt{s26\_m1\_t1\_tug\_data.json} and\\
\texttt{s26\_m1\_t1\_bd\_results.csv}.

\begin{table}[htbp]
  \centering
  \caption{Selected metrics from the example run \texttt{s26\_m1\_t1}.}
  \label{tab:example_run_metrics}
  \begin{tabular}{ll}
    \toprule
    \textbf{Metric} & \textbf{Value} \\
    \midrule
    Turn direction & Right \\
    Total TUG time & 26.93 s \\
    Stand phase duration & 1.48 s \\
    First gait duration & 5.81 s \\
    Turn duration & 2.79 s \\
    Second gait duration & 5.76 s \\
    Sit phase duration & 3.85 s \\
    Cadence & 67.91 steps/min \\
    Velocity & 0.347 m/s \\
    Steps (first gait/second gait) & 12 / 11 \\
    XCoM deviation (first gait/second gait) & 0.027 / 0.041 m \\
    \bottomrule
  \end{tabular}
\end{table}

Because LaTeX does not natively render HTML pages as inline figures, the HTML reports are
treated as supplementary digital artifacts in this version of the manuscript. They preserve
interactive visual inspection while the paper reports the quantitative summary.

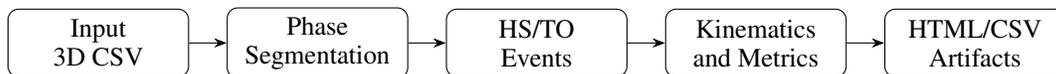
\begin{figure}[htbp]
  \centering
  \begin{tikzpicture}[
    node distance=1.0cm and 0.5cm,
    box/.style={draw, rounded corners, align=center, minimum width=2.4cm, minimum height=0.9cm},
    >={Stealth}
  ]
    \node[box] (inputCsv) {Input\\3D CSV};
    \node[box, right=of inputCsv] (phaseSeg) {Phase\\Segmentation};
    \node[box, right=of phaseSeg] (gaitEvents) {HS/TO\\Events};
    \node[box, right=of gaitEvents] (kinMetrics) {Kinematics\\and Metrics};
    \node[box, right=of kinMetrics] (outputs) {HTML/CSV\\Artifacts};

    \draw[->] (inputCsv) -- (phaseSeg);
    \draw[->] (phaseSeg) -- (gaitEvents);
    \draw[->] (gaitEvents) -- (kinMetrics);
    \draw[->] (kinMetrics) -- (outputs);
  \end{tikzpicture}
  \caption{High-level \textit{tugturn.py} processing flow used in this study.}
  \label{fig:pipeline}
\end{figure}

\section{Discussion}
\textit{tugturn.py} contributes a targeted software layer for TUG-specific markerless analysis, emphasizing phase-aware processing rather than generic full-trial summaries. This focus is important for protocols where turning, sit-to-stand transitions, and return gait carry distinct clinical meaning.

From a software-engineering perspective, the combination of CLI execution, TOML configuration, and structured outputs facilitates reproducibility, batch scaling, and downstream statistical workflows. The report stack (HTML plus CSV) supports both exploratory interpretation and scripted analysis.

Current limitations include dependence on upstream pose-estimation quality, protocol-specific threshold tuning, and the absence of benchmarked agreement against instrumented laboratory gold standards in this manuscript. Future work should include cross-dataset validation, uncertainty quantification, and multicenter reproducibility studies.

\section{Conclusion}
This work presents \textit{tugturn.py} as a reproducible and practical software pipeline for markerless 3D TUG analysis. By combining spatial phase segmentation, phase-filtered gait events, Vector Coding, and XCoM metrics in a single workflow, \textit{tugturn.py} enables consistent extraction of clinically relevant biomechanical descriptors from routine video-derived data. The provided test workflow and command-line interface support transparent reuse and adaptation in biomechanics and rehabilitation research.

\section*{Availability}
The software, test files, and execution examples used in this manuscript are available in the project repository. The \texttt{tugturn.py} example dataset and TOML configuration support direct reproducibility of the documented command-line runs.

\appendix
\section{Example Run Reports in HTML}
The complete result reports for the reference run (\texttt{s26\_m1\_t1}) are provided as HTML artifacts in
\texttt{tugturn/results/}. These files are the primary run-level result outputs generated by the pipeline:

\begin{itemize}
  \item \texttt{tugturn/results/s26\_m1\_t1\_tug\_report.html}
  \item \texttt{tugturn/results/s26\_m1\_t1\_tug\_report\_interactive.html}
\end{itemize}

Additional machine-readable outputs from the same run are:
\begin{itemize}
  \item \texttt{s26\_m1\_t1\_tug\_data.json}
  \item \texttt{s26\_m1\_t1\_bd\_results.csv}
  \item \texttt{s26\_m1\_t1\_bd\_steps.csv}
  \item \texttt{s26\_m1\_t1\_bd\_kinematics.csv}
  \item \texttt{s26\_m1\_t1\_bd\_vector\_coding.csv}
  \item \texttt{s26\_m1\_t1\_bd\_participants.csv}
\end{itemize}

In this manuscript, quantitative summaries are reported in the main Results section, while the HTML reports
are retained in the appendix as full visual and interactive evidence of the execution output.

The full PDF report exported from this run is included below as appendix pages, preserving the complete
clinical output view used for disability-focused cohorts (including Parkinson's disease):
\texttt{images/TUG Report\_ s26\_m1\_t1.pdf}.



\section*{Supplementary material}

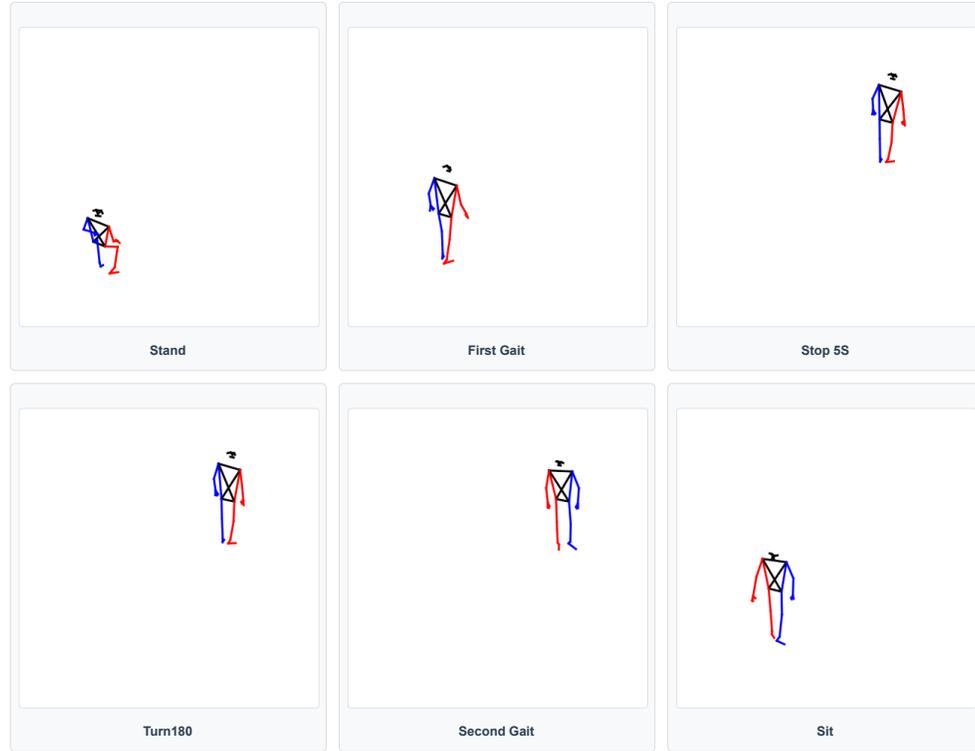
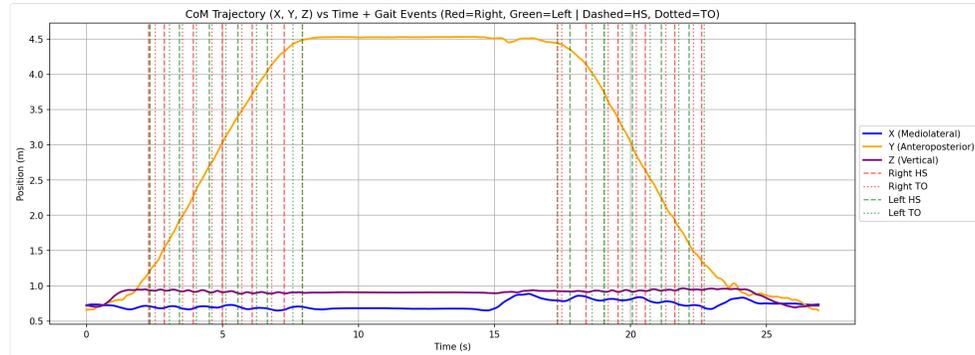
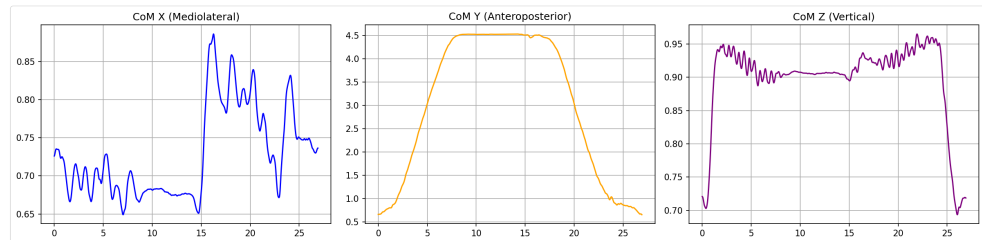

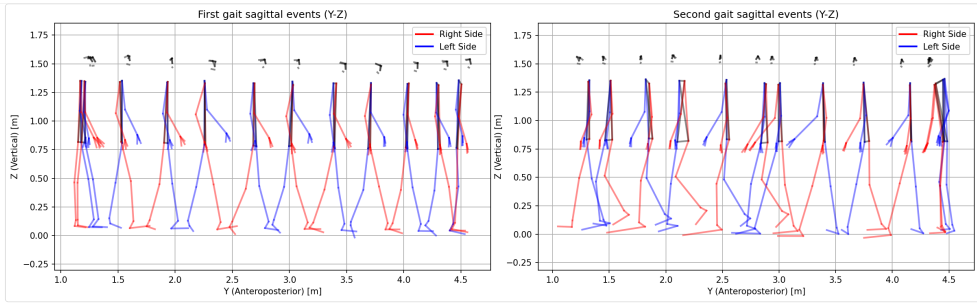

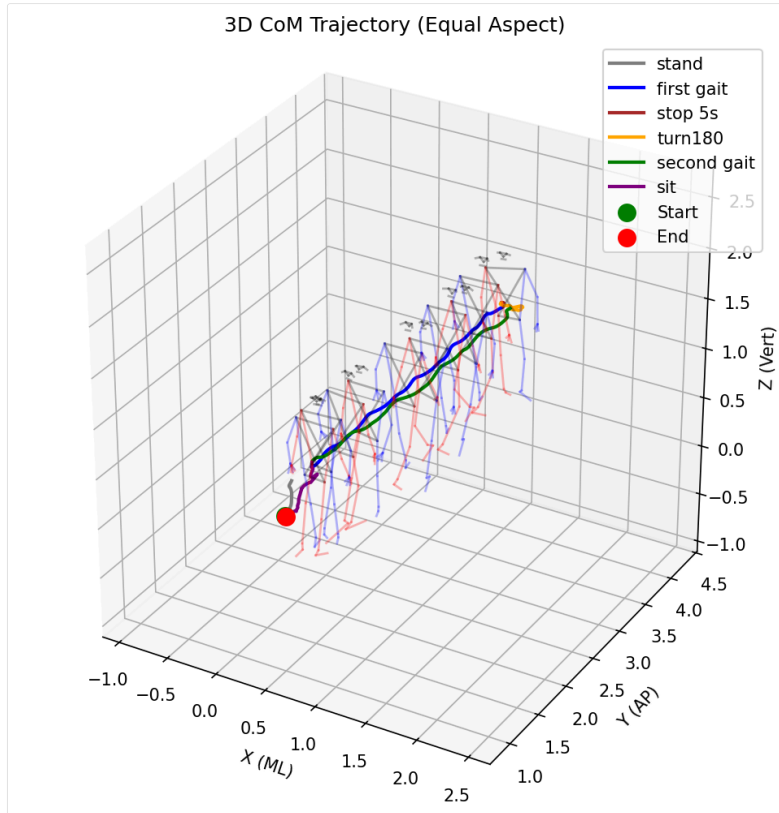

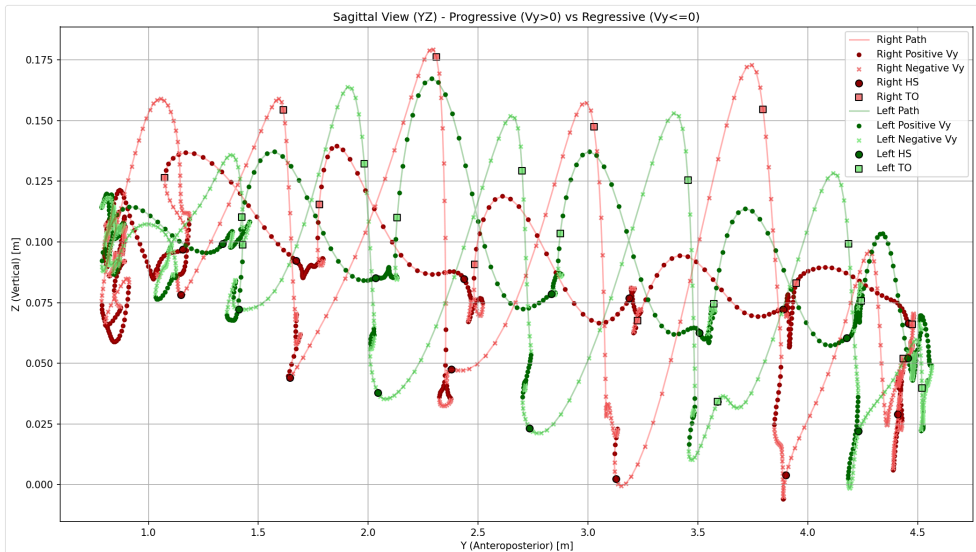

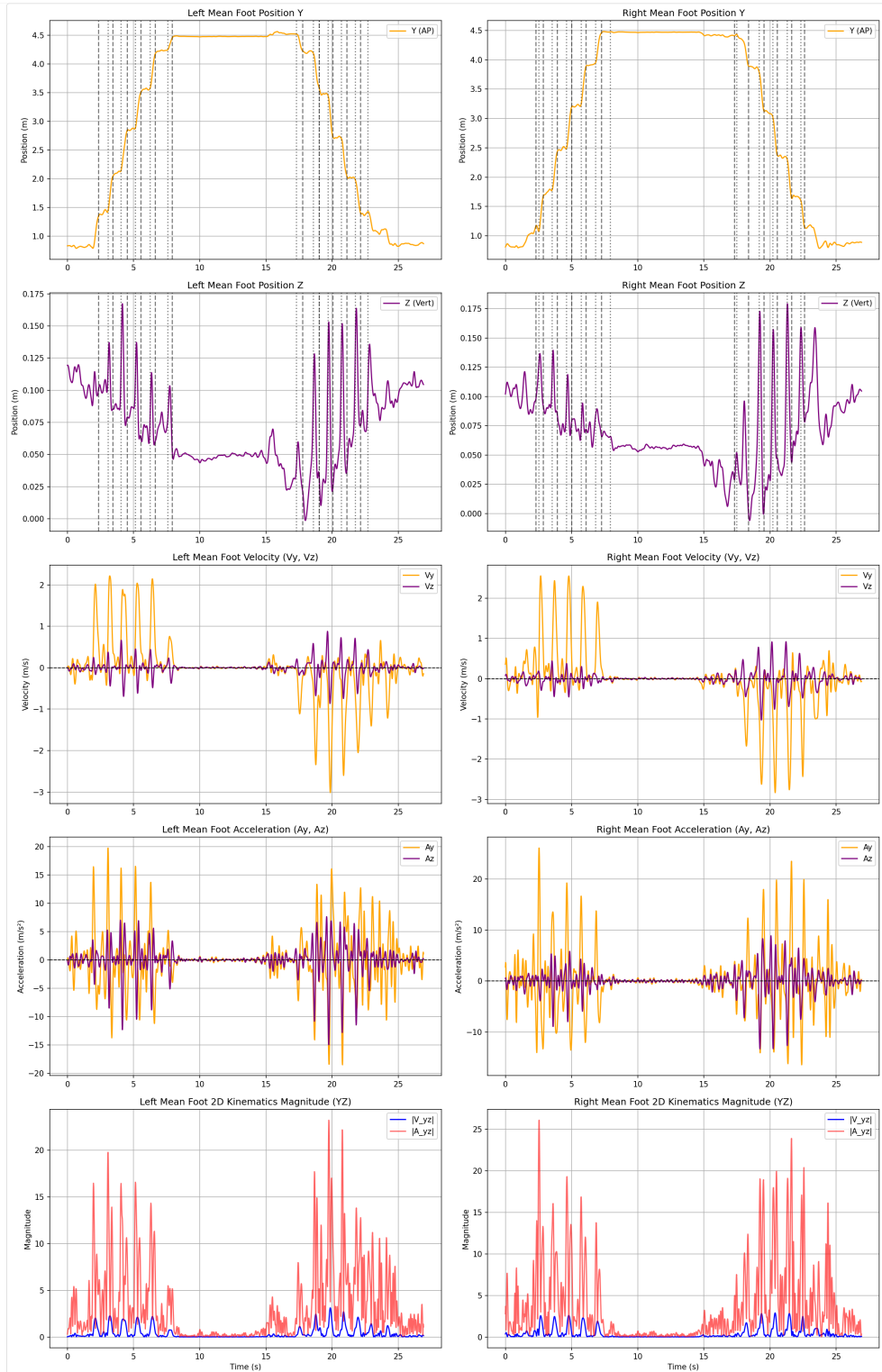

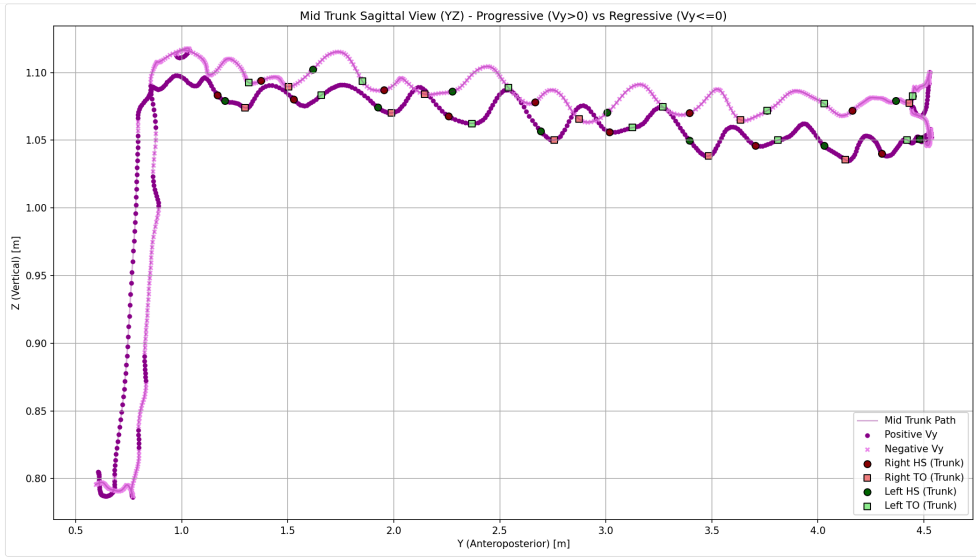

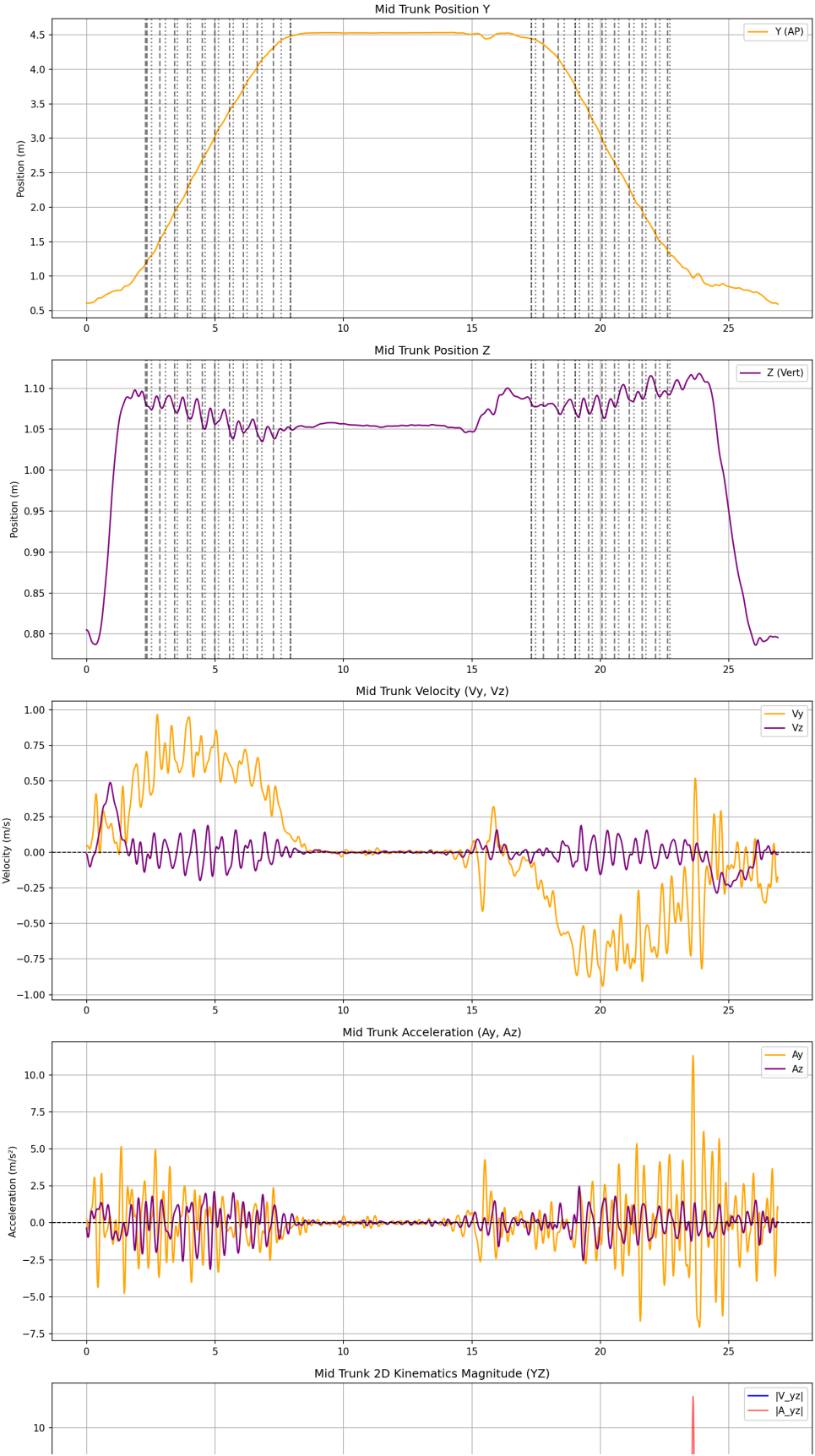

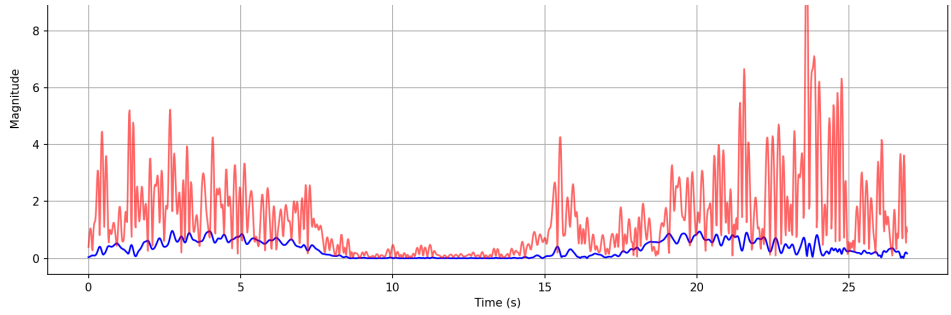

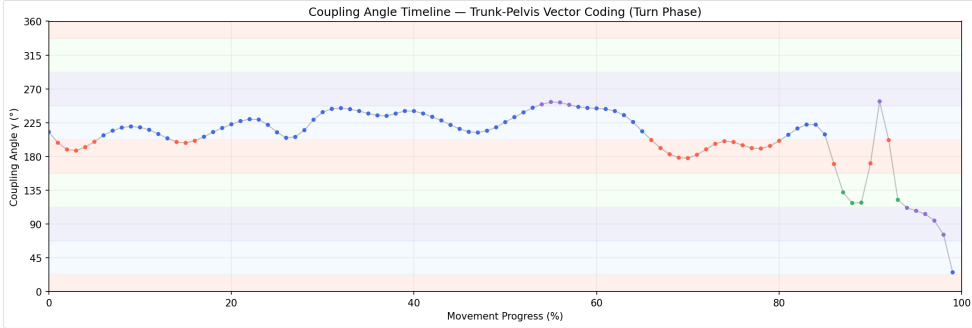

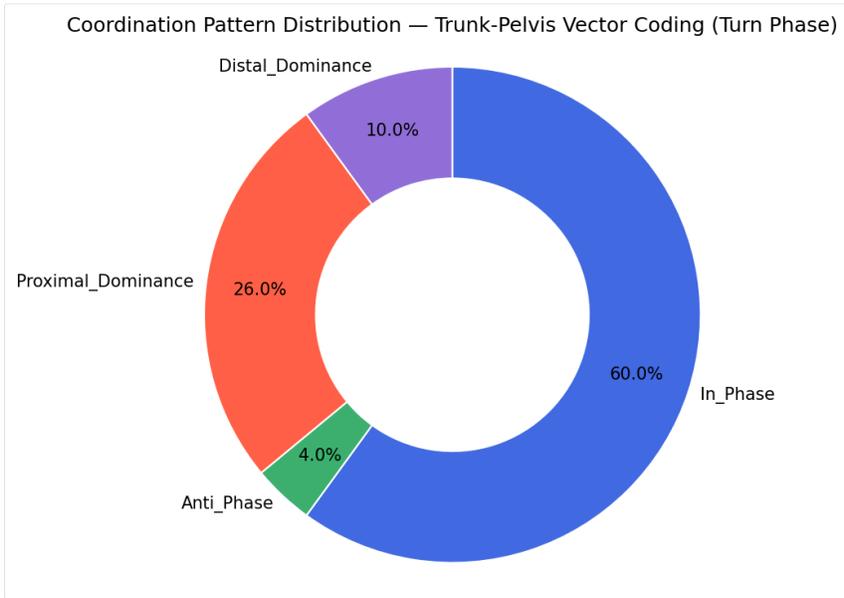

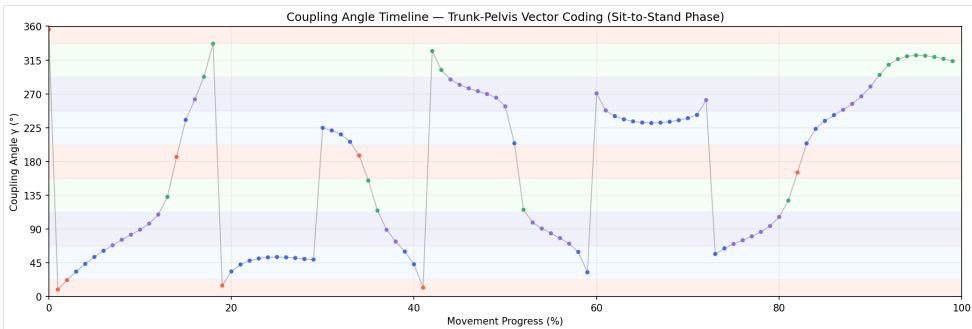

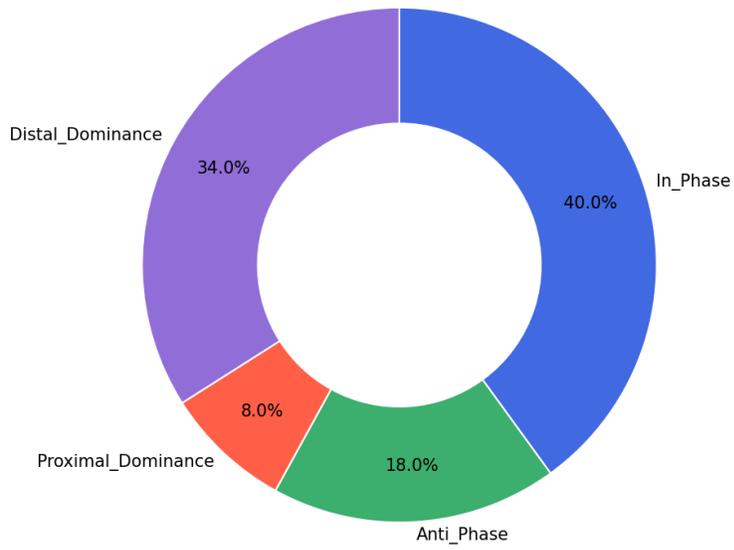

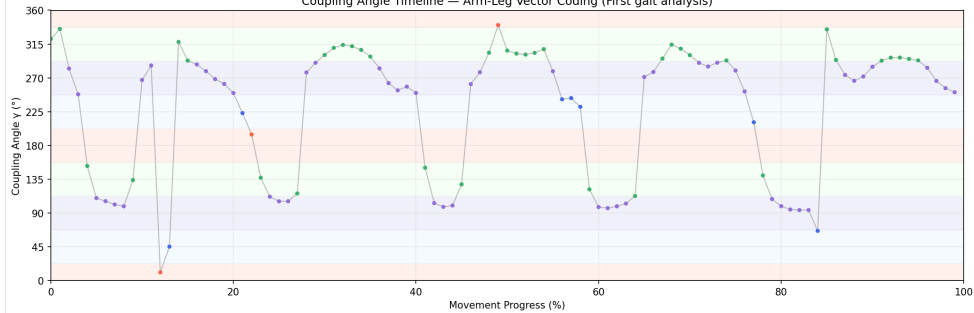

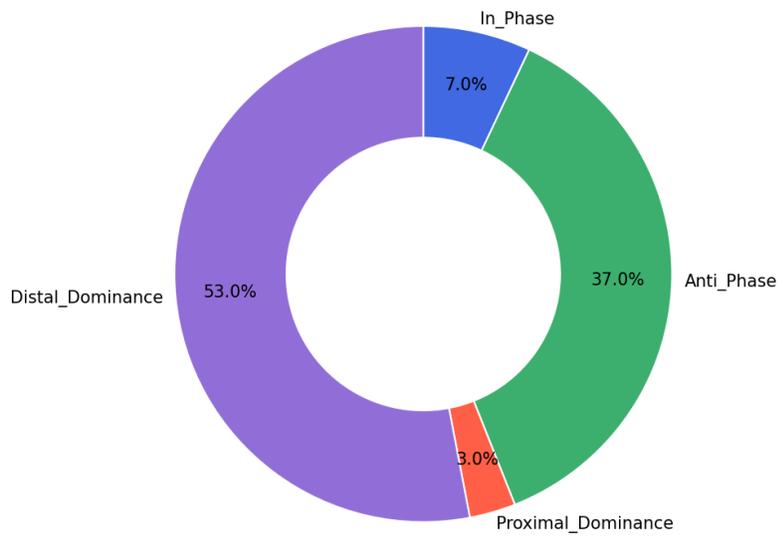

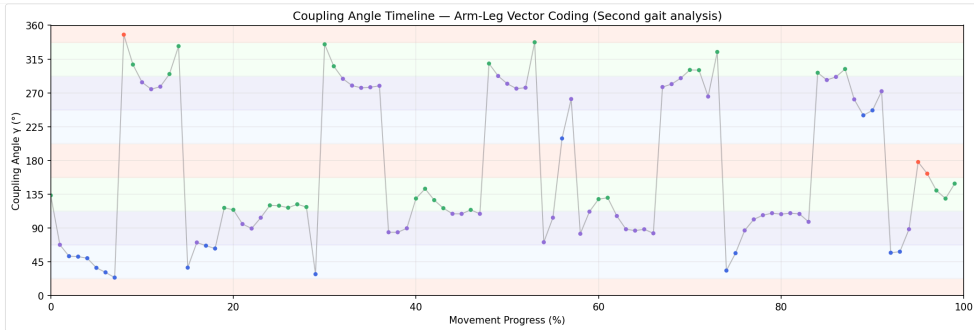

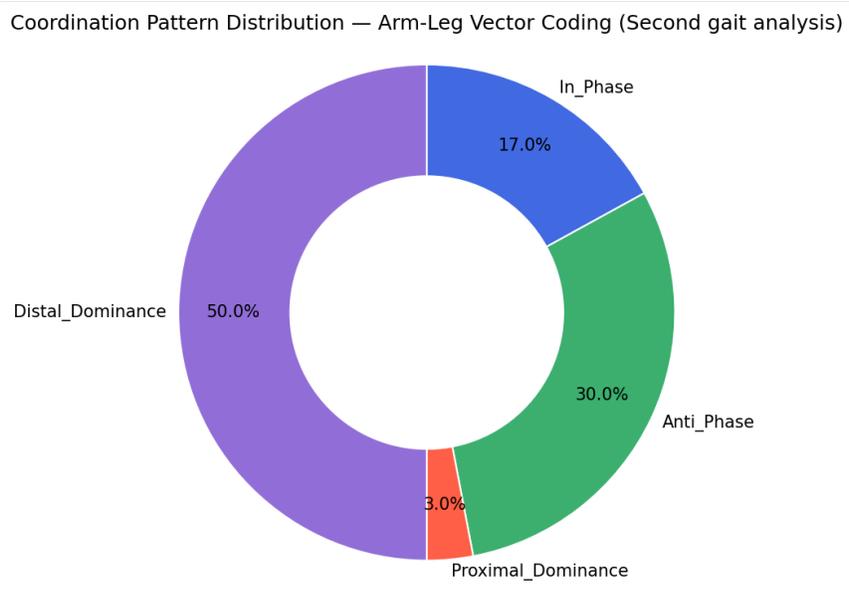

**Metadata**

| Field | Value |
| --- | --- |
| **SEX** | M |
| **WEIGHT** | 87.3 |
| **HEIGHT** | 1.63 |
| **AGE** | 63.0 |
| **FPS** | 59.94005994005994 |
| **STAGE_HY** | 3.0 |
| **PERFORMANCE** | 4 |
| **TURNSIDE** | R |

**Spatiotemporal Parameters**

**Global**

| Variable | Value |
| --- | --- |
| **Cadence steps per min** | 67.912 |
| **Velocity m s** | 0.347 |
| **XcoM Deviation First Gait m** | 0.027 |
| **XcoM Deviation Second Gait m** | 0.041 |
| **Steps First Gait** | 12.000 |
| **Steps Second Gait** | 11.000 |
| **First 1st strike inside zone** | R |

| Variable | Value |
|---|---|
| **First 1st complete step inside** | L |
| **First Number of steps inside** | 11.000 |
| **First Last strike inside zone** | L |
| **First Sequence steps inside** | ['R', 'L', 'R', 'L', 'R', 'L', 'R', 'L', 'R', 'L'] |
| **Second 1st strike inside zone** | R |
| **Second 1st complete step inside** | L |
| **Second Number of steps inside** | 10.000 |
| **Second Last strike inside zone** | R |
| **Second Sequence steps inside** | ['R', 'L', 'R', 'L', 'R', 'L', 'R', 'L', 'R', 'L', 'R'] |

**Left vs Right**

| Variable | Left | Right |
|---|---|---|
| **Stance Time s** | 0.691 | 0.694 |
| **Stance Time sd** | 0.104 | 0.029 |
| **Step Length m** | 0.321 | 0.368 |
| **Step Length sd** | 0.096 | 0.038 |
| **Step Width m** | 0.264 | 0.264 |
| **Step Width sd** | 0.077 | 0.063 |
| **Stride Length m** | 0.690 | 0.658 |
| **Stride Length sd** | 0.043 | 0.111 |
| **Swing Time s** | 0.409 | 0.365 |
| **Swing Time sd** | 0.040 | 0.043 |

**Step-by-Step Timeseries (Gait Events / HS)**

| Step Index | Side | Time (s) | Y (m) | Phase | Stance Time (s) | Step Length (m) | Step Width (m) | Stride Length (m) | Swing Time (s) |
|---|---|---|---|---|---|---|---|---|---|
| 1 | Right | 2.286 | 1.155 | First Gait | 0.2336 | 0.5956 | 0.1528 | 0.4829 | 0.3337 |
| 2 | Left | 2.336 | 1.337 | First Gait | 0.7174 | 0.1816 | 0.1575 | 0.686 | 0.367 |
| 3 | Right | 2.853 | 1.671 | First Gait | 0.6673 | 0.3394 | 0.1905 | 0.7657 | 0.4004 |
| 4 | Left | 3.420 | 2.034 | First Gait | 0.6173 | 0.3877 | 0.2439 | 0.8206 | 0.4671 |
| 5 | Right | 3.921 | 2.435 | First Gait | 0.684 | 0.4186 | 0.2611 | 0.7642 | 0.3837 |
| 6 | Left | 4.505 | 2.835 | First Gait | 0.6173 | 0.4243 | 0.3158 | 0.6699 | 0.4338 |
| 7 | Right | 4.988 | 3.187 | First Gait | 0.7174 | 0.3612 | 0.2916 | 0.7095 | 0.3837 |
| 8 | Left | 5.556 | 3.507 | First Gait | 0.684 | 0.3382 | 0.2903 | 0.6746 | 0.4004 |
| 9 | Right | 6.089 | 3.887 | First Gait | 0.7174 | 0.3886 | 0.3068 | 0.5643 | 0.4505 |
| 10 | Left | 6.640 | 4.178 | First Gait | 0.9343 | 0.3031 | 0.2651 | 0.2732 | 0.3504 |
| 11 | Right | 7.257 | 4.458 | First Gait | 0.6673 | 0.2973 | 0.2315 | 0.0618 | 9.376 |
| 12 | Left | 7.925 | 4.456 | First Gait | 9.376 | 0.1202 | 0.1066 | 0.3125 | 0.4671 |
| 13 | Right | 17.301 | 4.409 | Second Gait | 0.1668 | 0.138 | 0.1511 | 0.4787 | 0.8842 |
| 14 | Left | 17.768 | 4.23 | Second Gait | 0.8175 | 0.252 | 0.2024 | 0.6636 | 1.0344 |
| 15 | Right | 18.352 | 3.901 | Second Gait | 0.8175 | 0.3179 | 0.31 | 0.7745 | 0.367 |
| 16 | Left | 19.019 | 3.59 | Second Gait | 0.6673 | 0.3882 | 0.3423 | 0.7806 | 0.367 |
| 17 | Right | 19.536 | 3.127 | Second Gait | 0.6673 | 0.4157 | 0.3243 | 0.7066 | 0.3337 |
| 18 | Left | 20.053 | 2.734 | Second Gait | 0.6507 | 0.4221 | 0.3611 | 0.7134 | 0.4171 |
| 19 | Right | 20.537 | 2.377 | Second Gait | 0.7508 | 0.3749 | 0.3233 | 0.7374 | 0.3337 |
| 20 | Left | 21.121 | 2.043 | Second Gait | 0.634 | 0.4 | 0.3364 | 0.6386 | 0.3837 |
| 21 | Right | 21.622 | 1.643 | Second Gait | 0.684 | 0.3907 | 0.3236 | 0.5956 | 0.3003 |
| 22 | Left | 22.139 | 1.411 | Second Gait | 0.5672 | 0.3137 | 0.2791 | | |
| 23 | Right | 22.606 | 1.148 | Second Gait | | 0.3742 | 0.3009 | | |

*tugturn* Automated Markerless 3D TUG                                                   A PREPRINT**TUGTURN Phases (Seconds)**

| Phase | Range |
|---|---|
| stand | 0.0s to 1.48s (Dur: 1.48s) |
| first gait | 2.17s to 7.97s (Dur: 5.81s) |
| stop 5s | 7.97s to 14.51s (Dur: 6.54s) |
| turn180 | 14.51s to 17.3s (Dur: 2.79s) |
| second gait | 17.3s to 23.06s (Dur: 5.76s) |
| sit | 23.06s to 26.91s (Dur: 3.85s) |
| Total TUG Time | 26.926900000000003 |
| Turn Direction | Right |

15



\bibliographystyle{unsrtnat}
\bibliography{references}

\end{document}